\documentclass[review, authoryear]{elsarticle}
\pdfoutput=1
\usepackage{lineno,hyperref}
\usepackage[final]{changes}
\usepackage{amsmath}
\usepackage{amssymb}
\usepackage{threeparttable}
\usepackage{booktabs}
\usepackage{makecell}
\usepackage{multirow}
\usepackage{subfigure}
\usepackage{bbding}
\usepackage{algorithm}
\usepackage{algpseudocode}  
\usepackage{newfloat}
\usepackage{listings}
\journal{Journal of \LaTeX\ Templates}

\biboptions{authoryear}

\begin{document}

\begin{frontmatter}

\title{MSAT: Biologically Inspired Multi-Stage Adaptive Threshold for Conversion of Spiking Neural Networks\tnoteref{mytitlenote}}
\tnotetext[mytitlenote]{Xiang He and Yang Li contributed equally to this work.}

\author[1,2,3]{Xiang He}
\ead{hexiang2021@ia.ac.cn}
\author[1,2,3]{Yang Li}
\ead{liyang2019@ia.ac.cn}
\author[1,3]{Dongcheng Zhao}
\ead{zhaodongcheng2016@ia.ac.cn}
\author[1,2,3]{Qingqun Kong\corref{mycorrespondingauthor}}
\ead{qingqun.kong@ia.ac.cn}
\author[1,2,3,4]{Yi Zeng\corref{mycorrespondingauthor}}
\ead{yi.zeng@ia.ac.cn}
\cortext[mycorrespondingauthor]{Corresponding author}

\address[1]{Institute of Automation, Chinese Academy of Sciences (CAS), BeiJing, China}
\address[2]{School of Artificial Intelligence, University of Chinese Academy of Sciences, BeiJing, China}
\address[3]{Brain-inspired Cognitive Intelligence Lab, Institute of Automation, Chinese Academy of Sciences, BeiJing, China}
\address[4]{Center for Excellence in Brain Science and Intelligence Technology, Chinese Academy of Sciences, Shanghai, China}

\begin{abstract}
  Spiking Neural Networks (SNNs) can do inference with low power consumption due to their spike sparsity. ANN-SNN conversion is an efficient way to achieve deep SNNs by converting well-trained Artificial Neural Networks (ANNs). 
  However, the existing methods commonly use constant threshold for conversion, which prevents neurons from rapidly delivering spikes to deeper layers and causes high time delay.
  In addition, the same response for different inputs may result in information loss during the information transmission. 
  Inspired by the biological model mechanism, we propose a multi-stage adaptive threshold (MSAT). Specifically, for each neuron, the dynamic threshold varies with firing history and input properties and is positively correlated with the average membrane potential and negatively correlated with the rate of depolarization. The self-adaptation to membrane potential and input allows a timely adjustment of the threshold to fire spike faster and transmit more information.
  Moreover, we analyze the Spikes of Inactivated Neurons error which is pervasive in early time steps and propose spike confidence accordingly as a measurement of confidence about the neurons that correctly deliver spikes. We use such spike confidence in early time steps to determine whether to elicit spike to alleviate this error.
  Combined with the proposed method, we examine the performance on non-trivial datasets CIFAR-10, CIFAR-100, and ImageNet. We also conduct sentiment classification and speech recognition experiments on the IDBM and Google speech commands datasets respectively. Experiments show near-lossless and lower latency ANN-SNN conversion. To the best of our knowledge, this is the first time to build a biologically inspired multi-stage adaptive threshold for converted SNN, with comparable performance to state-of-the-art methods while improving energy efficiency.  
\end{abstract}


\end{frontmatter}

\section{Introduction}

	Artificial Neural Networks (ANNs) have been widely used in speech recognition, image processing, and other fields. However, with network structures becoming more complex, they often require large amounts of computational resources. In addition, the current ANNs computing paradigm differs from the human brain's mechanism, which communicates by transmitting spike trains generated by action potentials. Spiking Neuron networks (SNNs) work similarly to the brains and transmit the spike sequence to the downstream neurons. The information of spikes is enormous, and their distribution is sparse, so SNNs could demonstrate low power consumption characteristics.  

	SNNs have the potential for efficient inference when combined with neuromorphic hardware. Furthermore, they inherently show efficiency in processing temporal and spatial data. Their diverse coding mechanisms and event-driven characteristics are also promising. However, because the transmitted spikes are not differentiable, it is challenging to train SNNs. To solve this problem, some algorithms based on surrogate gradients (SG) \citep{lee2016training, wu2018spatio, wu2019direct, zhang2020temporal, shen2022backpropagation} and spike-timing-dependent plasticity (STDP) \citep{caporale2008spike, diehl2015unsupervised, hao2020biologically, zhao2020glsnn} have been proposed. However, it is still difficult to train deeper SNNs with complex network structures from scratch, resulting in a massive gap in performance between SNNs and ANNs in recognition and detection tasks.

	To bridge the performance gap between SNNs and ANNs, methods of converting well-trained ANNs to SNNs with the same structure have been proposed. The fundamental principle supporting such conversion is that firing rates of spiking neurons could approximate their counterparts activation (ReLU) in ANNs with sufficient time steps \citep{diehl2015fast}.

  In most existing conversion schemes, the thresholds of neurons in the same layer are identical and equal to the maximum activation value of that layer in the ANN. For this mechanism, spatially, the neuron needs to accumulate more membrane potential to exceed a pre-set threshold to deliver the spike, which makes the deeper neuron need to wait longer to receive the spike from the prior layer, and temporally, the neuron delivers the same information when firing the spike at different time steps, regardless of how much the membrane potential exceeds the threshold, causing an information loss.
  While in the biological counterpart of the spiking neuron, the spiking thresholds exhibit large variability, which has a significant impact on the encoding of inputs to spiking neurons \citep{fontaine2014spike}. The threshold value is not fixed even for the same neuron and threshold variability can be considered as an inherent characteristic of biological neurons, which exhibits self-adaptation to the membrane potential over a short time scale \citep{azouz2000dynamic, henze2001action, azouz2003adaptive, pena2002postsynaptic, wilent2005stimulus}. Threshold variability also enhances coincidence detection properties of cortical neurons \citep{azouz2000dynamic, wilent2005stimulus} as mentioned in 
\citep{fontaine2014spike}. Therefore, from the biological plausibility perspective, the different threshold across neurons and time steps is genuine feature of neurons and has a profound impact on the input coding.

  Motivated by the above biological model mechanisms, we propose a multi-stage adaptive threshold (MSAT) for ANN-SNN conversion to effectively take advantage of the spatial information brought by different neurons in the same layer and the temporal information brought by the accumulation of membrane potentials over time steps. For each neuron, the dynamic threshold is related to the firing history, i.e., positively correlated with the preceding membrane potential, and to the input characteristics, i.e., negatively correlated with the rate of depolarization. 
  In addition, we analyze the conversion error layer-to-layer and show that the Spikes of Inactivated Neurons (SIN), i.e., the misfired spikes which are due to the transient characteristics of the spikes, are pervasive in the early time steps and cause degradation of accuracy. According to statistical analysis of the error introduced by SIN in each layer, we propose spike confidence to measure the confidence about the correct neurons rather than misfired neurons and use spike confidence in early time steps to determine whether to elicit spike or not. 
  We conduct experiments on standard object classification benchmarks and non-visual domains such as natural language processing and speech to validate the effectiveness and universality of the proposed method.

Our major contributions can be summarized as follows:
\begin{itemize}
  \item We use the widely existing mechanism in the nervous system for reference and propose multi-stage adaptive threshold (MSAT) for converted SNNs for the first time. Different from constant threshold, self-adaptation to input value and membrane potential reduces latency and information loss.
  \item We formulate the layer-wise conversion error and derive the Spikes of Inactivated Neurons error. Accordingly, we propose spike confidence to alleviate SIN error. Each spike is given a confidence according to the statistical SIN error, and the spike confidence determines whether to fire a spike rather than fire directly.
  \item We conduct comprehensive experiments and ablation studies to show that the multi-stage adaptive threshold and spike confidence measurement facilitate faster transmission of information and reduce information loss in spike trains. Experimental results show that our proposed method has a comparable performance with SOTA methods but consumes less energy.
\end{itemize}

\section{Related work}

The conversion methods use the well-trained ANN and map it to an equivalent SNN. The studies begin with \citep{perez2013mapping} and the main idea that the mean firing rate in Integrate-and-Fire (IF) neuron \citep{burkitt2006review} can approximate ReLU activation value is proposed by \citep{diehl2015fast}. A mathematical statement of the feasibility of conversion is expressed firstly in \citep{rueckauer2017conversion}. Since \citep{sengupta2019going}, the converted SNNs begin going deeper and doing classification tasks in larger datasets. However, performance degradation is still the main problem. \citep{han2020rmp} propose a more efficient reset method named soft-reset, which uses a reset by subtraction mechanism. \citep{rathi2020diet} first use threshold optimization in deep SNN.
Although the performance of the converted SNN is improved, the cost of large time delays also comes.  \citep{deng2021optimal} divide conversion into floor and clip error from a new quantization perspective, \citep{li2021free} further optimize the conversion error.
\citep{yu2021constructing} construct the deep SNN with double-threshold, \citep{liu2022spikeconverter} propose temporal separation to further zip gap between ANN and SNN. \citep{li2022efficient} use burst mechanism and propose LIPooling to solve the conversion error caused by the MaxPooling layer.
\citep{wang2022towards} present a dual-phase converting algorithm to relieve clip and floor error. \citep{bu2021optimal, bu2022optimized} consider membrane potential initialization and quantization of the activation function to obtain faster SNN.
Nonetheless, most previous works have been limited to the fixed threshold and ignored utilizing the individual neurons to transmit information effectively. Some studies \citep{kim2020towards,li2021free} take threshold variation into consideration \added{while they still use the two-stage or heuristic method for SNNs, which require careful design or extensive search.}
\citep{ding2022biologically} use the biologically dynamic threshold mechanism to do robot obstacle avoidance and continuous control tasks and enhance the host SNN generalization. \added{In all, a bio-plausible multi-stage threshold for deep spiking neural networks has not been explored yet.}

\begin{table}[!t]
  \centering
  \begin{threeparttable}
    \resizebox{\linewidth}{!}{
      \begin{tabular}{cc | cc}
        \toprule
        Symbol & Definition &Symbol &Definitionp\\
        \midrule
        $l$ &Layer index & $M^l$ &Neuron numbers in layer $l$\\
        $i$ &Neuron index &$W$ &Weight connection\\
        $t$ &Current time step & $a$ & ANN activation value\\
        $T$ &Total time step & $\mathbf{V}_i^l(t)$ & Membrane potential before firing\\
        $V_{th}$ &Constant threshold &$V_i^l(t)$ & Membrane potential after firing\\
        $V_{th}(t)$ &Dynamic threshold & p & Spike confidence\\
        $\Theta_{t,i}^{l}$ &Step function indicating spike at t & $e$ & conversion error in total time step\\
        $c$ & Spike filter &E &Early time step\\
        \bottomrule
    \end{tabular}
    }
  \end{threeparttable}
  \caption{Summary of notations in this paper.}
  \label{notations}
\end{table}

\section{Conversion errors analysis}
In this section, we first review the neuron models of ANNs and SNNs. Then we analyze the conversion errors with constant threshold and adaptive threshold separately. The symbols used in this paper are summarized in Table \ref{notations}.
\subsection{Neuron model}
In ANN, the activation value $a_i^l$ (after ReLU) of neuron $i$ in layer $l$ can be computed as
\begin{equation}
  a_i^l = \operatorname{max}\left(0, \sum_{j=1}^{M^{l-1}}W_{ij}^la_j^{l-1} + b_i^l\right),
\end{equation}
where $l \in \{1, \cdots ,L\}$ indicates layer $l$ in a network with L layers; $W_{ij}^l$ is the weight connection between neuron $i$ in layer $l$ and neuron $j$ in layer $l-1$;
$b_i^l$ means neuron $i$ bias in layer $l$ and is constant all the time; we omit the bias for convenience in the following description. The number of neurons in layer $l$ is $M^l$, and activation value $a_i^l$ starts from $l=0$ and $a^0=x$ for the direct input.

The membrane potential before firing at time step $t$, $\mathbf{V}_i^l(t)$ is a sum of the last time step $t-1$ membrane potential and current input. When $\mathbf{V}_i^l(t)$ exceeds a certain voltage threshold, it emits an output spike and resets the membrane potential.
One of the most widely adopted models is Integrate-and-Fire (IF) neuron, and the membrane potential would then be updated by soft-reset mechanism, which subtracts the threshold in $\mathbf{V}_i^l(t)$ rather than reset the membrane potential to $V_{\text{reset}}$. We use $V_i^l(t)$ to denote membrane potential after firing, then the mathematical form is as follows 
\begin{equation}
	\label{eq2}
	V_i^l(t)=V_{i}^{l}(t-1)+V_{th}^{l-1}\sum_j^{M^{l-1}}W_{ij}^l\Theta_{t,j}^{l-1}-V_{th}^l\Theta_{t,i}^{l},
\end{equation}
where $\Theta_{t,i}^{l}$ is a function indicating the neuron $i$ in layer $l$ elicits a spike at time $t$. Here we use $\Theta_{t,i}^l$ to replace $\Theta_{i}^l(t)$ for simplicity.
\begin{gather}
  \Theta_{t,i}^{l}=\mathcal{H}\left(V_{i}^{l}(t-1)+z_{i}^{l}(t)-V_{th}^l\right) \notag,  \\
 \mathcal{H}(x)= \begin{cases}1, & \text { if } x \geq 0 \\ 0, & \text { else. }\end{cases}
\end{gather}
Here $z_i^l(t)$ is the input of neuron $i$ in layer $l$ and time $t$:
\begin{equation}
  z_{i}^{l}(t) = V_{th}^{l-1}\sum_j^{M^{l-1}}W_{ij}^l\Theta_{t,j}^{l-1}.
\end{equation}

\subsection{Constant threshold conversion error}
The main idea in ANN-SNN conversion is using mean firing rate $r_i^l(t)$, which indicates the firing rate of neuron $i$ in layer $l$ for a total time $t$, to approximate the activation value $a_i^l$.
Here we give an analytical explanation for the approximation process.

\begin{figure}[!t]
  \centering
  \includegraphics[width=1.0\textwidth]{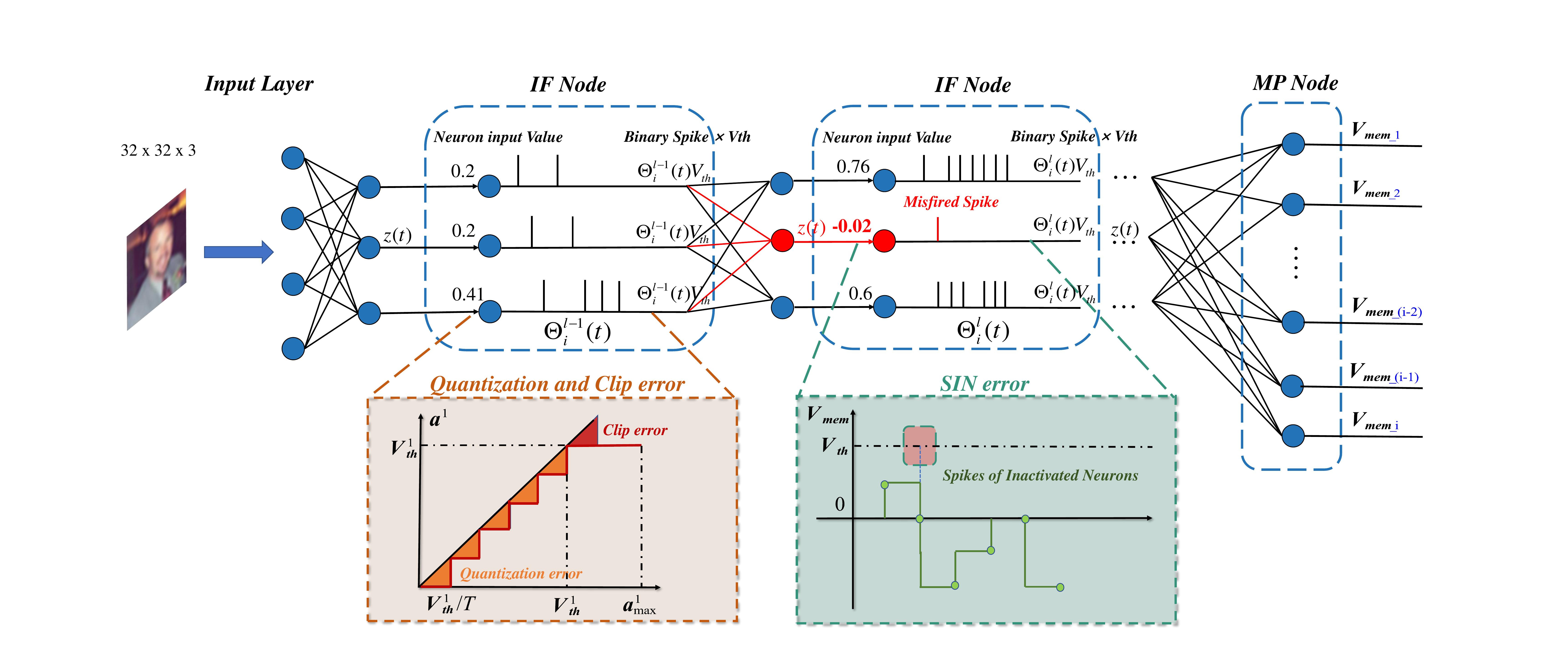}
  \caption{A multilayer perceptron spiking neural network within 10 time steps for demonstrating conversion error.
  The red box shows the quantization error and clip error, caused by the discrete and insufficient time steps.
  The green box shows the Spikes of Inactivated Neurons error, caused by the dynamic transients of the neurons. }
   \label{fig1}
\end{figure}

The conversion error comes from two-part: converting ANN to SNN directly, resulting in quantization error and clip error ${e}_{i,QC}^l$; the other is transient characteristic of neurons and irregular elicited spike, resulting in Spikes of Inactivated Neurons \citep{li2021bsnn} error ${e}_{i,SIN}^l$, as shown in Fig. \ref{fig1}. 
In the subsequent error analysis, we assume that the firing rate of the preceding layer fully represents the activation value, which means ${r}_i^{l-1}={a}_i^{l-1}$. Then the conversion error in layer $l$ could be the following equation:

\begin{equation}
	{e}_i^l = {r}_i^l - {a}_i^l = {e}_{i,QC}^l + {e}_{i,SIN}^l.
\end{equation}

\subsubsection{quantization and clip Error}

For Eq. \ref{eq2}, accumulating the input over the simulation time step $T$, we can derive the firing rate $r_i^l(T)$ relationship layer by layer,
\begin{equation}
  r_i^l(T) = \sum_j^{M^{l-1}}W_{ij}^lr_j^{l-1}(T) - \frac{V_i^l(T)}{T}.
  \label{eq:7}
\end{equation}

Here $r_i^l(T)=V_{th}^l\Theta_{t,i}^{l}/T$. When $V_{th}^l$ is larger than the maximum activation value, $V_i^l(T)$ will be less than $V_{th}^l$. Thus, the residual membrane potential cannot be output, that is why information transmission suffers a loss.
Because of the discreteness of the time steps, the mean firing rate can be described as a step floor function and cannot precisely approximate the continuous ReLU function, known as quantization error or flooring error. For example, as shown in Fig. \ref{fig1}, activation value 0.42 should be accurately represented by 42 spikes within 100 time steps, while 4.2 spikes cannot be achieved in discrete 10 time steps because the number of spikes should be an integer.

If the voltage threshold is set smaller than the maximum activation value, then when the $\mathbf{V}_i^l(t)$ exceeds the threshold, the emitted spike will not transmit effective information to distinguish membrane potential above the threshold, which is known as clip error.
Setting voltage threshold to maximum activation value can avoid this but suffers a notable latency.

Considering the clip operation, the actual firing rate can be expressed as
\begin{equation}
	\label{eq9}
  r_i^l(T) = \operatorname{clip}\left(\frac{V_{th}^l}{T}\left\lfloor\frac{\sum_j^{M^{l-1}}W_{ij}^lr_j^{l-1}T}{V_{th}^l}\right\rfloor, 0, V_{th}^l\right),
\end{equation}
where $r_j^{l}$ means $r_j^{l}(T)$ for simplifying the statement and $\sum_j^{M^{l-1}}W_{ij}^lr_j^{l-1}$ is $a_i^l$. Eq. \ref{eq9} shows that even assuming the same inputs in the previous layer, the average firing rate of the SNN is still differ from the activation values in the ANN due to discrete quantization and clip, thus generating quantization and clip errors ${e}_{i,QC}^l$.

\subsubsection{Spikes of Inactivated Neurons error}
Inactivated Neurons refer to the neurons whose activation value counterparts in ANN are negative. Theoretically, they should not fire spike in all time steps to achieve ReLU filtering the negative value. However, once they elicit spikes, the corresponding mean firing rate $r_i^l$ will be larger than zero. As shown in Fig. \ref{fig1}, the adjacent weights of the red neuron are respectively 0.5, 0.5, and -0.5, then the simulation activation value is -0.01, meaning that no spike should have been fired. However, a spike fired by mistake results in SIN error.

We use $\mathcal{R}=\left\{j \middle\vert\ \sum_{t=0}^T\Theta_{t,j}^l >0,\  a_j^l(t) < 0\right\}$ to denote neurons with SIN, then SIN error can be expressed as
\begin{equation}
  {e}_{i,SIN}^l = \mathbf{0} - \frac{\sum_{j \in \mathcal{R}} \sum_{t=0}^TW_{ij}^l\Theta_{t,j}^l }{T}.
\end{equation}

\begin{figure}[!t]
  \centering
  \includegraphics[width=0.95\columnwidth]{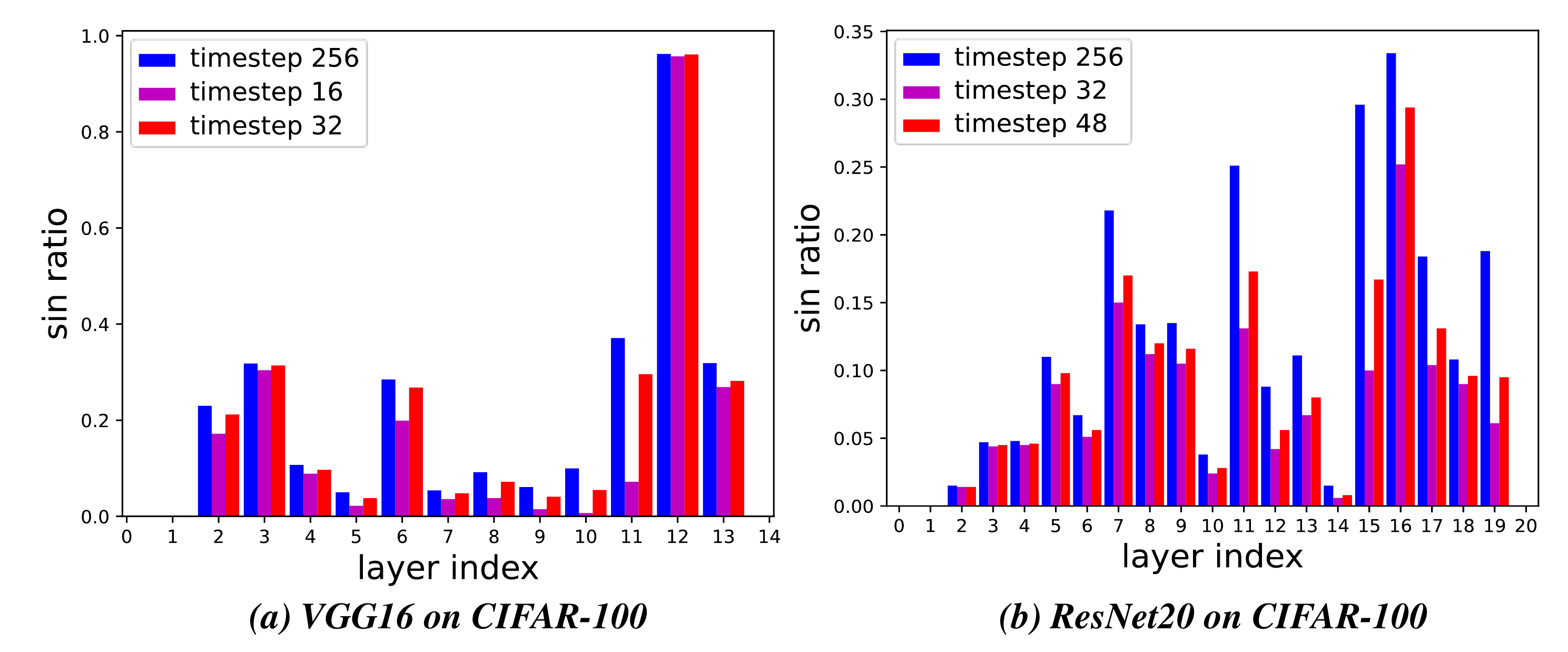}
  \caption{Neurons with SIN ratio in each layer of VGG16 and ResNet20 in CIFAR-100.}
  \label{fig:2}
\end{figure}

\added{By observing outputs of each neuron at different time steps,} we find that the neurons with SIN usually fire at an early stage and then keep silent at a later stage during the conversion process.
We count neurons with SIN proportion in each layer, as shown in Fig. \ref{fig:2}. It shows that neurons responsible for SIN are prevalent and take the larger proportion in the deeper layer.
Moreover, it is clear that most SIN appear in the early time steps. Since SIN mean undesired error existed in spike, thus SIN lead to inaccurate coding of signals and degradation of the accuracy.

\subsection{Adaptive threshold optimization error}
In this part we demonstrate that when the threshold is a function of time step $t$, spike trains are able to transmit the equivalent or more efficient information than the constant threshold.

With the threshold that changes over time steps, Eq. \ref{eq2} can be rewritten in the following form:

\begin{equation}
  V_i^l(t)=V_{i}^{l}(t-1)+\sum_j^{M^{l-1}}V_{th,j}^{l-1}(t)W_{ij}^l\Theta_{t,j}^{l-1} -V_{th,i}^l(t)\Theta_{t,i}^{l}.
\end{equation}

The firing rate during time step T, $r_i^l(T)$, is computed as $\sum_{t'=1}^{T}V_{th,i}^l(t')\Theta_{t',i}^{l}/T$, which means Eq. \ref{eq:7}, the firing rate relationship in the higher layer still exists.
On this basis, the residual neuron membrane potential $V_i^l(T)$ can be appropriately adjusted, so the spike information could be more efficient and thus shorten the conversion latency. The optimization target in threshold adaptation is 
\begin{gather}
    \min\limits_{V_{th,i}^l(T)}\left(\mathbf{o} - \mathbf{o^\prime} \right), \notag \\
    \mathbf{o} = \operatorname{clipfloor}\left(\sum_j^{M^{l-1}}W_{ij}^lr_j^{l-1}, T, V_{th,i}^l(T)\right)\\ - \frac{\sum_{j \in \mathcal{R}} \sum_{t=0}^TW_{ij}^l\Theta_{t,j}^l }{T}, \notag \\
    \mathbf{o^\prime} = \operatorname{ReLU}\left(\sum_j^{M^{l-1}}W_{ij}^lr_j^{l-1}\right). \notag
\end{gather}

There is no closed-form solution to the above problem, \citet{li2021free} use gird search to heuristically find the final solution.
A trivial solution is to make $V_{th,i}^l=\sum_j^{M^{l-1}}W_{ij}^lr_j^{l-1}$, which means the voltage threshold equals to input value for each neuron. With this solution, ANN can be converted to SNN only one time step. However, such SNN elicits spike every time step and lose spike sparsity, so it is unreasonable and makes no sense, thus should not be used.
In the next section, we will give a biologically rational method.

\section{Methods}
\begin{figure}[!t]
  \centering
  \subfigure[MSAT schematic diagram]{
    \includegraphics[width=4.5in]{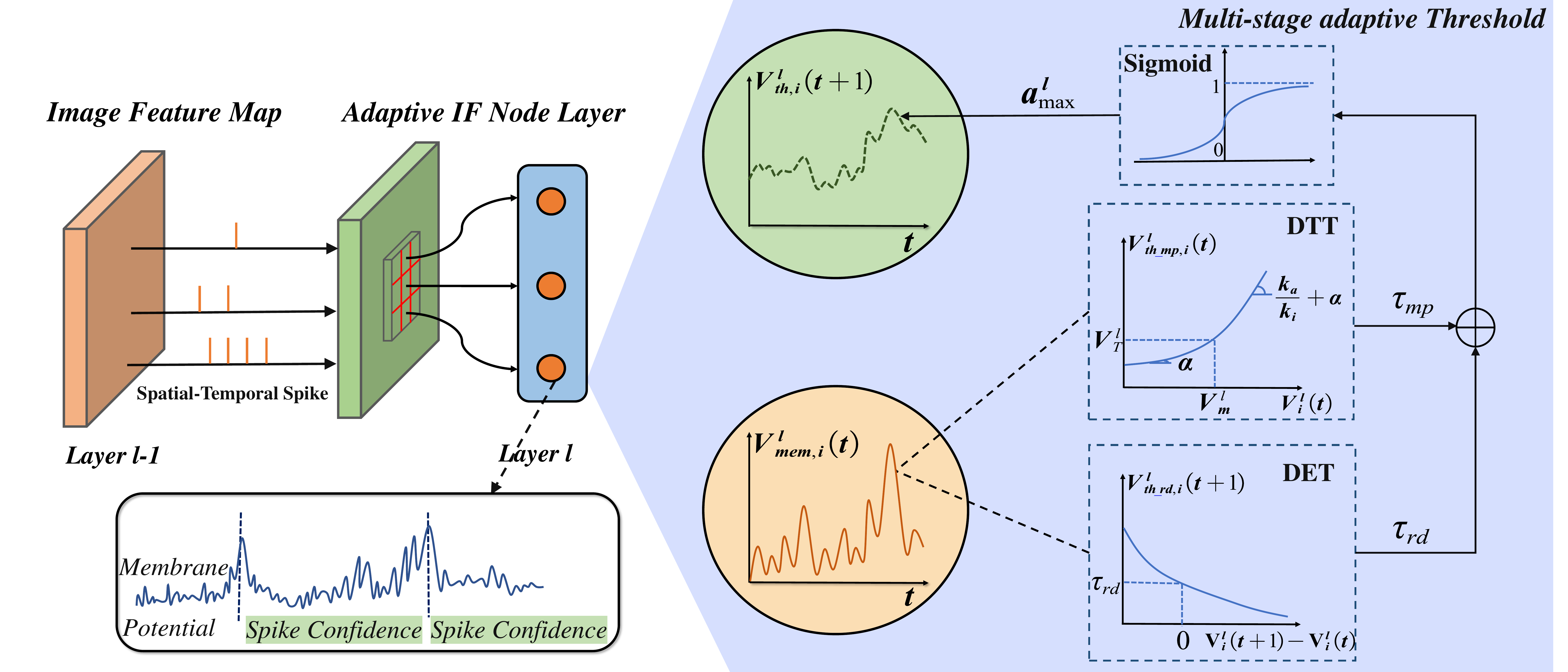}
    \label{fig:3a}
  }
  \subfigure[Spike confidence]{
    \includegraphics[width=3.5in]{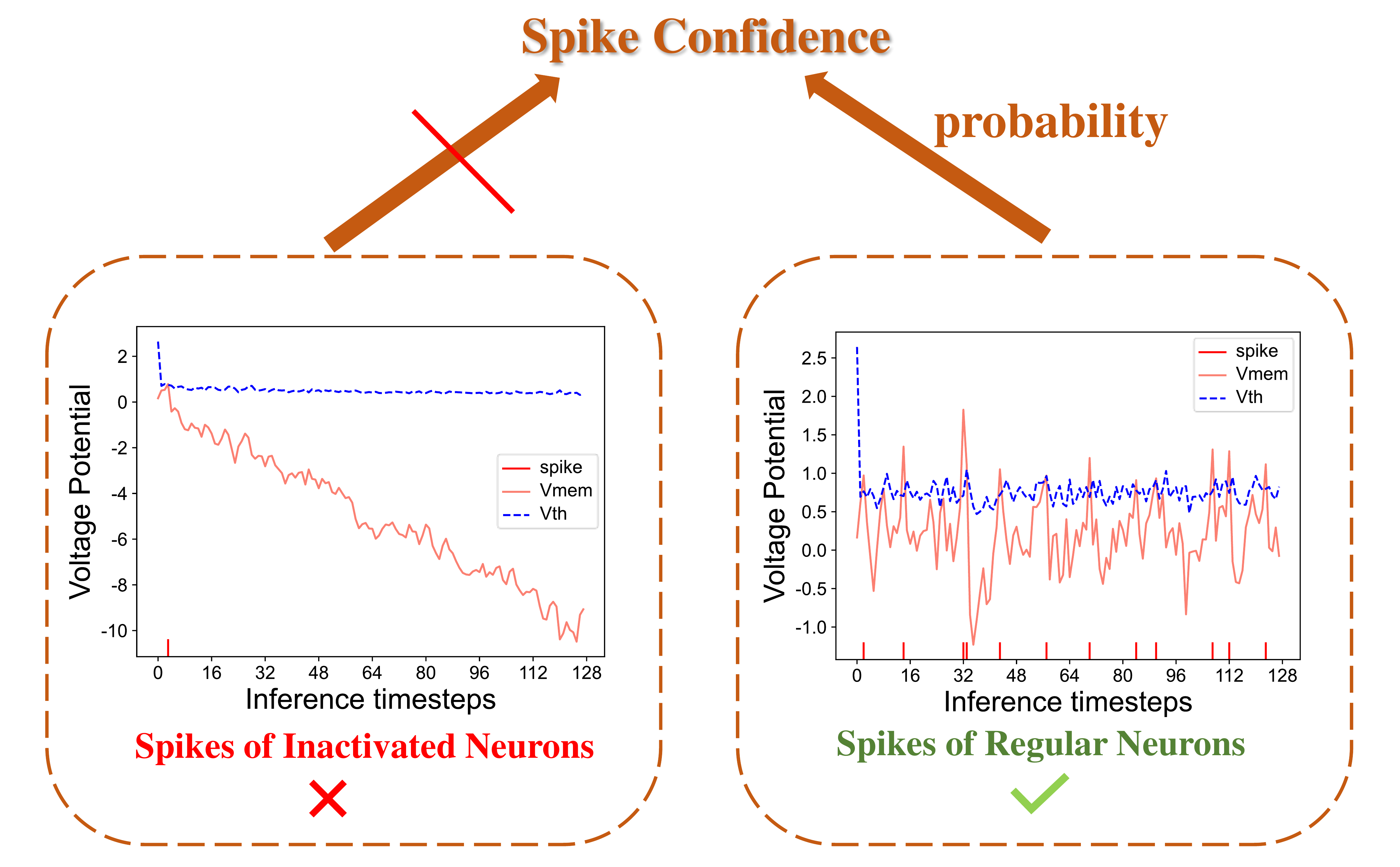}
    \label{fig:3b}
  }
  \label{fig:3}
  \caption{Summary of MSAT and spike confidence. Spike confidence is only implemented in the early stages.}
\end{figure}

\subsection{Multi-Stage Adaptive Threshold}
\deleted{Some other threshold adaptation models, such as the threshold increases after each spike and decreases if there is no spike\citep{querlioz2013immunity, diehl2015unsupervised} are relatively simple and not fitted to this problem, so we do not adopted them. }
Based on the above analysis, the adaptive threshold, which is multi-stage and varies with inference time, could better facilitate information from individual neurons and different time steps.
For each neuron, the adaptive threshold varies with firing history and input properties.
The adaptive threshold at time step $t+1$, $V_{th,i}^l(t+1)$ can be described as
\begin{equation}
  V_{th,i}^l(t+1) = \tau_{mp}V_{th\_mp,i}^l(t) + \tau_{rd}V_{th\_rd,i}^l(t+1),
\end{equation}
where $\tau_{mp}$ and $\tau_{rd}$ are the time constant of the dynamic tracking threshold $V_{th\_mp,i}^l(t)$ and dynamic evoked threshold $V_{th\_rd,i}^l(t+1)$ separately. Specifically, the dynamic tracking threshold is positively correlated with the average preceding membrane potential and the dynamic evoked threshold is negatively correlated with the rate of depolarization. 

Fig. \ref{fig:3a} gives the multi-stage adaptive threshold schematic diagram: when a neuron receives current input, its threshold will take average membrane potential and rate of depolarization into consideration, then the threshold will vary with these two factors. Finally, it gives a changed threshold as the adaptation to membrane potential and rate of depolarization. 

\subsubsection{Dynamic Tracking Threshold} 
Dynamic Tracking Threshold (DTT) is a reflection of spiking threshold that varies with firing history. 
In \citet{fontaine2014spike}, the DTT is a similar first-order kinetic equation; here we use the steady-state threshold for fitting our SNNs.
Let us use $V_{m,i}^l(t)$ to denote the average membrane potential at time step $t$ in layer $l$ neuron $i$, then DTT can be described as follows
\begin{equation}
  V_{th\_mp,i}^l(t) = \alpha\left(V_i^l(t)-V_{m,i}^l(t)\right)+V_{T}^l+k_aln\left(1+e^{\frac{V_i^l(t)-V_{m,i}^l(t)}{k_i}}\right),
  \label{eq:15}
\end{equation}
where $\alpha, k_a, k_i$ are all time constant, $V_{T}^l$ is the parameter to optimize. When residual membrane potential $V_i^l(t)$ is less than the average membrane potential $V_{m,i}^l(t)$, the slope of the curve around the inflection point is $\alpha$ and $\frac{k_a}{k_i} + \alpha$ respectively. $\alpha, k_a, k_i, V_{T}^l, V_i^l(t)$, these parameters together determine the curvature.

Eq. \ref{eq:15} shows that when the residual membrane potential increases, the threshold correspondingly increases and vice versa. There are several benefits that come along with such DTT scheme; for example, small voltage fluctuation generated by a small current input will have no effect on the output spike if it is smaller than the amplitude of the threshold adaption. In the early spike unstable phase, the neuron receives the synaptic current and adjusts the threshold according to the residual membrane potential, and if it receives a positive total input current but has not yet delivered a spike, the threshold is raised according to Eq. \ref{eq:15}, thus alleviating the Spikes of Inactivated Neurons error.

\subsubsection{Dynamic Evoked Threshold} 
Dynamic Evoked Threshold (DET) is a reflection of spiking threshold that varies with input properties, \citet{azouz2000dynamic} shows voltage threshold varies inversely with the preceding rate of depolarization $dV_m/dt$ by plotting scatter. 
 In IF neurons, we use variation of membrane potential before firing to express the preceding rate
 of depolarization thus DET can be expressed as
\begin{equation}
  V_{th\_rd,i}^l(t+1) = \tau_{rd}e^{-\frac{\left(\mathbf{V}_i^l(t+1)-\mathbf{V}_i^l(t)\right)}{C}}.
  \label{eq:DET}
\end{equation}

As mentioned before, here $\mathbf{V}_i^l(t)$ denotes membrane potential before firing and makes a distinction with $V_i^l(t)$, which denote residual membrane potential after firing.
$C$ is the time constant representing the sensitivity to input membrane potential variation. The threshold decreases with input value exponentially.

\subsubsection{MSAT with DET and DTT} 

MSAT shows that threshold-adapted neurons are insensitive to slow changes and selective to fast input variations with DET and DTT. In other words, adaptive thresholds filter out slow voltage fluctuations and corresponding neurons will not elicit a spike. The slow voltage fluctuations may come from the unexpected Spikes of Inactivated Neurons, so it relieves the SIN error partly and reduces the total spike number \replaced{thus}{meanwhile} promoting energy efficiency. On the other hand, the threshold variation by DET makes threshold not increases endlessly and reduces appropriately for fear of large quantization error.

\begin{algorithm}[!th] 
	\caption{Conversion from ANN to SNN: Multi-stage adaptive threshold} 
  \textbf{Input}: Pretrained ANN with L layer, training set\\
  \textbf{Parameter}: inference time step T, spike confidence p, early time step E, threshold coefficient Cof\\
  \textbf{Output}: The multi-stage adaptive SNN
  \begin{algorithmic}[1] 
    \For {s = 1 to \# of samples}
      \State $a^l \gets$ layer-wise activation value
      \For {$l$ = 1 to L}
        \State $V_{th}^l \gets max[V_{th}^l, max(a^l)]$
      \EndFor
    \EndFor

    \For {t = 1 to T}
      \For {$l$ = 1 to L}
        \State spike filter $c^l \gets Bernoulli(p[l])$
        \For{$j$ = 1 to neuron number of layer $l$}
          \State compute DTT and DET as Eq. \ref{eq:15} Eq. \ref{eq:DET}
          \State Cof$[l][j] \leftarrow  sigmoid(DTT+DET)$
          \State SNN.layer$[l].V_{th}[j]\gets$ Cof$[l][j] * V_{th}^l$
          \If{fire spike and t $<$ E}
          \State spike = $c^l$ * spike
          \EndIf
        \EndFor
      \EndFor
    \EndFor

	\end{algorithmic} 
  \label{alg:1}
\end{algorithm}

\subsection{Spike confidence}
As shown in Fig. \ref{fig:2}, the early spikes elicited are not always reliable, and some parts of them are raised by SIN. Although threshold adaptive DTT can partly relieve this, neurons with SIN will not be distinguished from other normal neurons until they keep silent in longer time step.
Inspired by many image recognition and detection works \citep{liu2016ssd, redmon2016you}, we import confidence to show how confident the elicited spikes are from the normal neurons to distinguish with SIN, as Fig. \ref{fig:3b} shows.
\added{Within early time steps E, the spike confidence acts and every spike which should elicit use this spike confidence to generate a spike filter to determine fire spike or not rather than fire directly. The value of E is determined based on the statistics of the SIN ratio.} 
Spike Confidence can be formulated as follows
\begin{gather}
  c_i^l \sim Bernoulli(p^l)
\end{gather}
where $p^l$ is spike confidence of neurons in layer $l$ and the value of $p^l$ eaquals to the opposite proportion of SIN in layer $l$.
Random variable $c_i^l$, which is sampled from the Bernoulli distribution whose parameter is defined by the spike confidence, determines whether to fire a spike instead of firing directly. In this case the accumulation of membrane potential becomes
\begin{gather}
\tilde{\Theta}_{t,i}^l = c_i^l * \Theta_{t,i}^l, \notag\\
V_i^l(t)=V_{i}^{l}(t-1)+\sum_j^{M^{l-1}}V_{th,j}^l(t)W_{ij}^l\tilde{\Theta}_{t,j}^{l-1} -V_{th,i}^l(t)\tilde{\Theta}_{t,i}^{l}.
\end{gather}

The MSAT and spike confidence ensure that all neurons adjust their thresholds and elicit spike according to the stimulus and in this way they become specialized.
The pseudocodes of adaptive threshold are shown in Algorithm \ref{alg:1}.

\section{Experiment}
Here we conduct experiments on CIFAR-10 \citep{krizhevsky2009learning}, CIFAR-100 \citep{krizhevsky2009learning}, and ImageNet \citep{russakovsky2015imagenet} datasets to demonstrate the effectiveness of our proposed method.
We use data augmentation, such as random horizontal flip, Cutout \citep{devries2017improved}, and AutoAugment \citep{cubuk2019autoaugment}. 
Our batch size is 128 and the total training epoch is 300. The optimizer chooses stochastic gradient descent (SGD) with an initial learning rate of 0.1 and uses a cosine decay strategy.
VGG16\citep{simonyan2014very}, ResNet20 and ResNet34\citep{he2016deep} are used for target ANN as in previous works for comparison.  Our code implementation of deep SNNs is based on the open-source SNN framework BrainCog \citep{zeng2022braincog}.

In the proposed adaptive thresholds, there are six hyperparameters and we give their meanings and values used in this paper in Table \ref{hyperparameters}. 
Although hyperparameters are manually set, they do not vary with the dataset and setting these parameters is not difficult based on our experiments; our method works well for a wide value range of these parameters. 
Although these parameters can be set to trainable, we choose to manually set them for the consideration of biological plausibility according to \citep{fontaine2014spike,azouz2000dynamic}.

\begin{table}[!t]
  \centering
  \resizebox{4.0in}{!}{
      \begin{tabular}{c c c  c c}
        \toprule
        Symbol& Definition &VGG16 &ResNet20 &ResNet34 \\
        \midrule
        $\alpha$ & left side slope & 0.03 & 0.3 & 1.0\\
        $k_a$ & right side slope hyperparameter& 1 & 1 &1.0\\
        $k_i$ & right side slope hyperparameter& 1.0 & 1.0 & 1.0\\
        $C$ &Input sensitivity & 5.0 & 5.0 & 5.0\\
        $\tau_{mp}$ &coefficient of DTT & 1 & 0.5 & 0.5\\
        $\tau_{rd}$ &coefficient of DET & 1 & 0.5 & 0.5\\
        \bottomrule
    \end{tabular}}
  \caption{Summary of hyperparameters on different network.}
  \label{hyperparameters}
\end{table}

\subsection{Comparison with the state of the art}
As mentioned in related work, some works train a modified ANN to get short latency when converting ANN to SNN. In order not to do too many restrictions on origin ANN, we choose to save the topology and use the original weights in the target ANN for universal conversion without retraining ANN. In this way, we compare the state-of-the-art approaches including TSC \citep{han2020deep}, RMP-SNN \citep{han2020rmp}, Opt \citep{deng2021optimal}, Calibration \citep{li2021free}, Burst \citep{li2022efficient}. 

As shown in Table \ref{cifar10andcifar-100}, with the proposed method, all conversions are achieved with minimal accuracy degradation and shorter latency. In particular, on CIFAR-10, VGG16, we achieved a lossless conversion in only 112 time steps. 
We also validated the robustness of our method on ImageNet. As shown in Table \ref{Res}, we also achieve comparative performance with SOTA methods and nearly lossless conversion from the target ANN. It is worth noting that though calibration \citep{li2021free} also uses the dynamic threshold, our method does not need to search for the best voltage threshold, which could result in less computational cost.

\begin{table}[!t]
  \centering
  \begin{threeparttable}
    \resizebox{\linewidth}{!}{
      \begin{tabular}{cccccc|ccccc}
        \toprule
        \multirow{2}{*}{Method} & \multirow{2}{*}{Use DT} &ANN &SNN &Loss &Step &ANN &SNN &Loss &Step\\
        \cmidrule(lr){3-6} \cmidrule(lr){7-10}
        & & \multicolumn{4}{c}{\textbf{VGG16, CIFAR-10}} & \multicolumn{4}{c}{\textbf{ResNet20, CIFAR-10}}\\
        \midrule
        RMP-SNN\citep{han2020rmp} &\XSolid&93.63\% & 93.63\% &$<$0.01\% & 2048 & 91.47\%& 91.36\% &0.11\% &2048\\
        TSC\citep{han2020deep} &\XSolid &93.63\% & 93.63\% &$<$0.01\% &2048 &91.47\% & 91.42\% &0.05\% &2048 \\
        Opt.\citep{deng2021optimal} &\XSolid& 92.34\% & 92.24\% &0.1\% & 128 & 93.61\% & 93.56\% &0.05\% &128\\
        Burst.\citep{li2022efficient} &\XSolid& 95.74\% & 95.75\% &0.02\% & 256 & 96.56\% & 96.59\% & -0.03\% &$<$256\\
        \textbf{Ours} &\Checkmark&95.45\% & 95.45\% &\textbf{0.00\%} & \textbf{112} &96.37\% & 96.36\% &\textbf{0.01\%} &\textbf{174}\\
        \midrule
        & & \multicolumn{4}{c}{\textbf{VGG16, CIFAR-100}} & \multicolumn{4}{c}{\textbf{ResNet20, CIFAR-100}}\\
        \midrule
        RMP-SNN\citep{han2020rmp} &\XSolid&71.22\% & 70.93\% &0.29\% & 2048 &68.72\% &67.82\% &0.9\% &2048\\
        TSC\citep{han2020deep} &\XSolid &71.22\% & 70.97\% &0.25\% & 2048 &68.72\% & 68.18\% &0.54\% &2048\\
        Opt.\citep{deng2021optimal} &\XSolid& 70.49\% & 70.47\% &0.02\% & 128 & 69.80\% & 69.49\% & 0.31\% &128\\
        Calibration\citep{li2021free} &\Checkmark& 77.89\% & 77.79\% & 0.1\% & $>$512 & 77.16\% & 77.29\% &-0.13\% & 64\\
        Burst\citep{li2022efficient} &\XSolid& 78.49\% & 78.66\% &-0.17\% & 128 & 80.69\% & 80.72\% &-0.03\% & $<$256 \\
        \textbf{Ours} &\Checkmark&78.49\% & 78.50\% &\textbf{-0.01\%} & \textbf{224} &80.69\% & 80.70\% &\textbf{-0.01\%} &\textbf{252}\\
        \bottomrule
    \end{tabular}
    }
  \end{threeparttable}
  \caption{Experimental results on CIFAR-10 and CIFAR-100, DT means dynamic threshold.}
  \label{cifar10andcifar-100}
\end{table}

\begin{table}[!t]
  \centering
  \begin{threeparttable}
    \resizebox{3.3in}{!}{
      \begin{tabular}{cccccc}
        \toprule
        Method & Use DT &ANN &SNN &Loss &Step\\
        \midrule
        RMP-SNN &\XSolid&70.64\% & 69.89\% &0.75\% & $>$2048\\
        TSC &\XSolid &70.64\% & 69.93\% &0.71\% &$>$2048\\
        Opt. &\XSolid&75.66\% & 75.44\% &0.22\% &$>$2048\\
        Calibration &\Checkmark&75.66\% & 75.45\% &0.21\% &$>$2048\\
        Burst. &\XSolid& 75.16\% & 74.94\% &0.22\% & 256\\
        \textbf{Ours} &\Checkmark&75.16\% & 74.93\% &0.23\% & 2045\\
        \bottomrule
    \end{tabular}
    }
  \end{threeparttable}
  \caption{Experimental results on ImageNet with ResNet34.}
  \label{Res}
\end{table}

\subsection{Performance on non-visual domains}
Our work aims to contribute a biologically adaptive threshold system for ANN-SNN conversion, not limited to computer vision tasks and so MSAT is equally competent in other tasks.
To show this, we further evaluate our method on non-visual domains, including natural language processing and speech.

For natural language processing, we choose IMDB dataset\citep{maas2011learning} for the sentiment classification task, and the network structure is three-layer bi-directional LSTM we built ourselves.
For the speech processing, we evaluate on the speech recognition task and select the google speech command dataset\citep{warden2018speech}. The dataset consisted of 105,829 utterances of 35 words and each stored as a one-second (or less) WAVE format file.
The specific architecture is modeled after the M5 network architecture described in \citet{dai2017very}.
We report the performance of MSAT on these two tasks and compare it with RMP-SNN. The results can be seen in Table \ref{nonvisual}. 

Compared to RMP-SNN, MSAT can achieve lossless conversion much faster, even in only half the time steps, e.g., 7 steps on IDBM dataset for LSTM. On complex tasks, such as speech recognition with 35 classes, MSAT performs better at the same time step. These results are consistent with earlier analysis and validate the effectiveness of MSAT.

\begin{table}[!t]
  \centering
  \begin{threeparttable}
    \resizebox{\linewidth}{!}{
      \begin{tabular}{ccccccc}
        \toprule
        \textbf{Task} & \textbf{Dataset} &\textbf{Network} &\textbf{ANN Accuracy}&\textbf{Method} &\textbf{Step} &\textbf{SNN accuracy}\\
        \midrule
        NLP: & \multirow{2}{*}{IDBM} &\multirow{2}{*}{LSTM} &\multirow{2}{*}{87.22\%} & RMP-SNN\footnotemark[1] & 11 &87.23\%\\
        \cmidrule(lr){5-7}
        sentiment classification& & & &MSAT &7 &87.25\% \\
        \midrule
        Speech:& \multirow{2}{*}{speech commands} &\multirow{2}{*}{M5} &\multirow{2}{*}{87.81\%} & RMP-SNN\footnotemark[1] & 2048 &86.83\%\\
        \cmidrule(lr){5-7}
        speech recognition& & & &MSAT &2048 &87.33\% \\
        \bottomrule
        \footnotetext[1] 0 1. Our implemented.
    \end{tabular}
    }
  \end{threeparttable}
  \caption{Experimental results on non-visual task.}
  \label{nonvisual}
\end{table}

\subsection{Ablation study}

\begin{figure}[!t]
  \centering
  \includegraphics[width=1.0\textwidth]{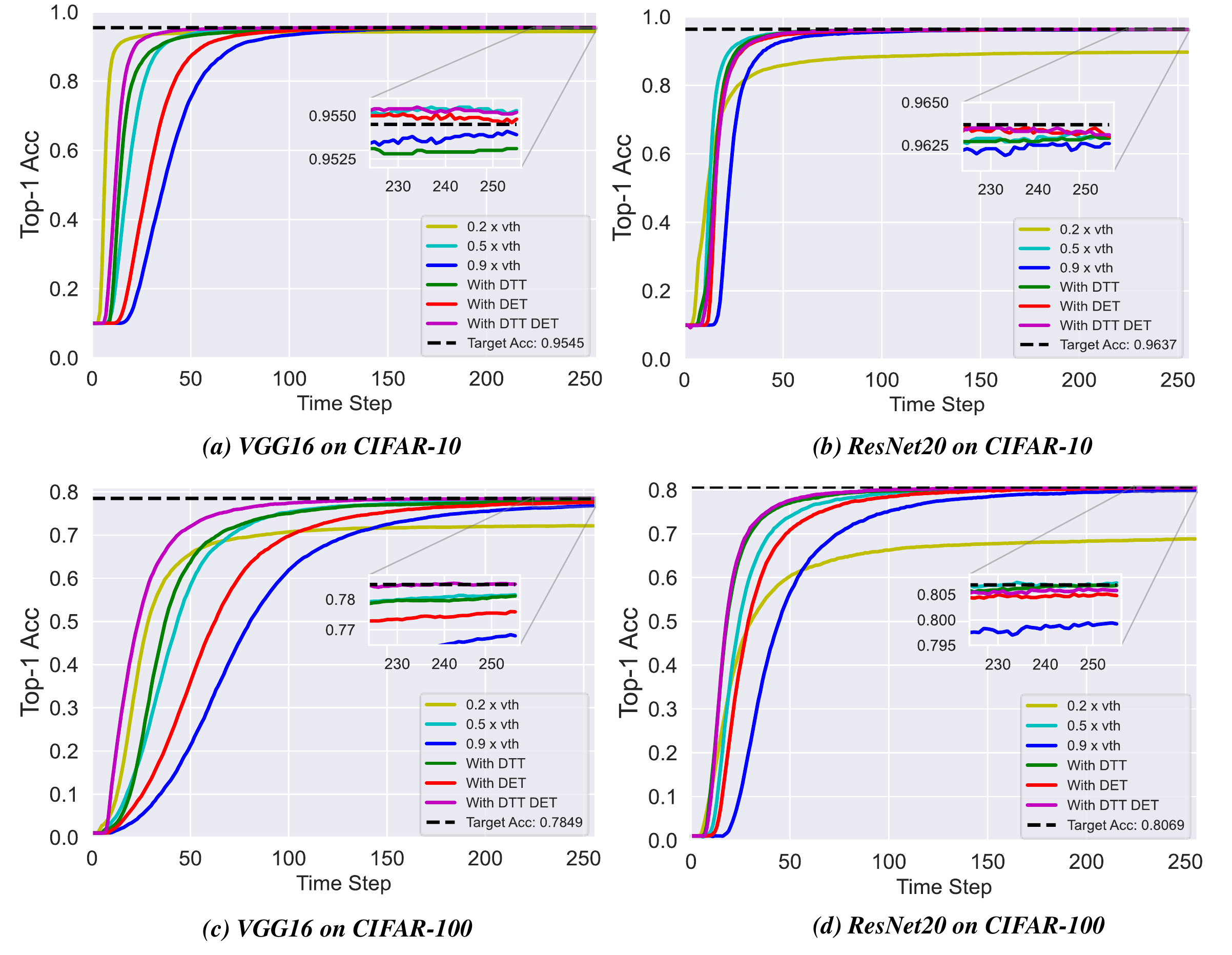}
  \caption{The accuracy curves on CIFAR-10 and CIFAR-100 datasets.}
  \label{fig:4}
\end{figure}

In our method, $V_{th}^l$ is chosen as the maximum activation value and as the initial value of the multi-stage adaptive threshold. In Fig. \ref{fig:4}, the dotted lines indicate the target ANN accuracy, from which we obverse the DTT, DET, and fixed threshold impact on classification accuracy.
It is intuitive that with either DTT or DET, we could achieve faster convergence rate, as shown in Fig. \ref{fig:4}(b), \ref{fig:4}(d).
Instead, the heuristic method is hard to achieve high accuracy and fast inference speed simultaneously. For example, $0.2V_{th}$ in Fig. \ref{fig:4}(a), \ref{fig:4}(b) (yellow lines) lie above other color lines in early stage and have a shorter latency; however, yellow lines do not appear in the zoomed-in box and can not touch dotted lines, which means that the converted SNN threshold with $0.2V_{th}$ suffers from larger performance degradation.
Although a manually selected threshold might achieve a satisfactory result, it needs lots of computation to search or carefully design for an appropriate threshold.
The adaptive threshold MSAT with both DTT and DET could have a faster inference to achieve the target ANN accuracy and shows that the MSAT could find an appropriate threshold at each time step to reduce the quantization error while not importing larger clip error.

Moreover, we validate the effect of spike confidence on accuracy in early time steps. In our experiment, we apply spike confidence to the last IF neuron layer because the last layer is directly responsible for calculating results. The spike confidence stage holds 16 time steps. As we mentioned before, SIN takes a large proportion, so a large part of the error comes from SIN besides quantization and clip error.
We use averaged number of SIN (ANS) over each neuron and the exclusion of spike confidence (SC) in both ANS and classification accuracy as baseline. Table \ref{Sc} shows that spike confidence could reduce the averaged number of SIN and improve classification accuracy, validating that SNN in the early time steps could elicit fewer error spikes and benefit from spike confidence.

\begin{table}[!t]
  \centering
  \begin{threeparttable}
    \resizebox{4.0in}{!}{
      \begin{tabular}{ccccc}
        \toprule
        \multirow{2}{*}{Metric} & \multicolumn{2}{c}{VGG16} &\multicolumn{2}{c}{ResNet20}  \\
        \cmidrule(lr){2-3} \cmidrule(lr){4-5}
        & CIFAR-10 & CIFAR-100 & CIFAR-10 & CIFAR-100 \\
        \midrule
        w/o SC ANS & $2.92$ & $4.467$ & $1.972$ & $6.309$ \\
        w/ SC ANS& $\mathbf{2.343}_{-0.577}$ & $\mathbf{2.514}_{-1.953}$ & $\mathbf{1.558}_{-0.414}$ & $\mathbf{3.063}_{-3.246}$\\
        w/o SC Acc (T=32) & $93.48$ & $61.16$  & $93.30$ & $71.92$ \\
        w/ SC Acc (T=32) & $\mathbf{93.85}_{+0.37}$ & $\mathbf{66.73}_{+5.57}$ & $\mathbf{94.02}_{+0.72}$ & $\mathbf{73.25}_{+1.33}$\\
        \bottomrule
    \end{tabular}
    }
  \end{threeparttable}
  \caption{Averaged Number of SIN and classification accuracy on different networks and datasets. The subscripts mean the decreased ANS and increased accuracy compared to baselines (1st and 3rd rows).}
  \label{Sc}
\end{table}

\begin{figure}[!t]
  \centering
  \includegraphics[width=1.0\textwidth]{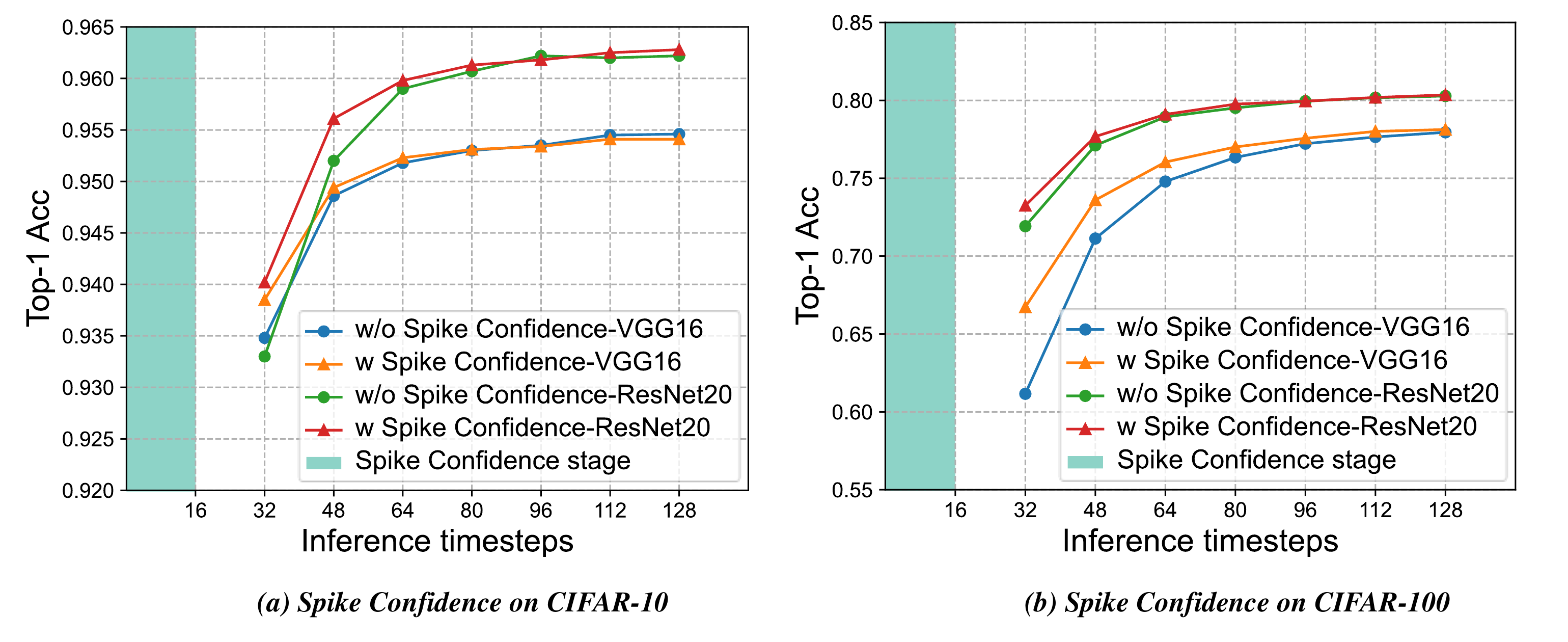}
  \caption{classification accuracy improves with spike confidence. Spike confidence is used within the time steps of the green area.}
  \label{fig:6}
\end{figure}

Fig. \ref{fig:6} demonstrates the impact of adding spike confidence on performance. The improvement on CIFAR-10 is not 
as significant as on CIFAR-100, mainly because the SIN proportion in CIFAR-10 is already quite smaller than that in CIFAR-100. To show this, we statistic the neurons with SIN ratio in the last layer for CIFAR-10-VGG16, CIFAR-10-ResNet20, CIFAR-100-VGG16 and CIFAR-100-ResNet20 respectively. The ratio is 0.059, 0.063, 0.617, and 0.506 respectively, which means that SIN error is a very small part in CIFAR-10 whose influence on performance degradation is not as significant as in CIFAR-100. This also explains the different degrees of performance improvement on the two datasets by spike confidence.

\subsection{Energy-efficiency and sparsity}
In this section, we compute the firing rate of each layer to evaluate our spike sparsity. We use the same way in \citet{rathi2020diet} to compute energy consumption.
32-bit floating-point AC and MAC per operation consume 0.9pJ and 4.6pJ individually.
Furthermore, there is no operation in SNN if no spike elicits and the first layer is the real-valued input so the energy consumption for this layer is still calculated as 4.6pJ per operation. We chose the VGG16 on CIFAR-100 dataset with time step T=64, the average firing rate is 0.0585 and we have 43.20\% energy consumption compared to the target ANN. This is more energy efficient than 69.30\% reported in \citet{li2022efficient}. Low firing rate and less than half the energy consumption reflects the spike sparsity of MSAT.

\section{Conclusion}
This paper introduces a multi-stage adaptive threshold into ANN-SNN conversion and demonstrates its advantages in terms of accuracy-latency trade-off. This is the first work to build a biologically inspired multi-stage adaptive threshold for converted SNN.
In addition, we focus on the SIN error that occupies a large proportion in the early period and accounts for accuracy degradation but is rarely explored in existing works. Through statistical analysis of SIN errors, we propose to mitigate SIN errors by using spike confidence.
Our experiments show that the converted SNN with MSAT and spike confidence has comparable accuracy to the state of the art while with lower latency and power consumption.
Good performance on classification tasks in non-visual domains also demonstrates the universality of the proposed approach.
% Use \bibliography{yourbibfile} instead or the References section will not appear in your paper

\section{Acknowledgments}
This study was supported by National Key Research and Development Program (Grant No.2020AAA0107800), the Strategic Priority
Research Program of Chinese Academy of Sciences (Grant No.XDB32070100), the Beijing Natural Science Foundation (Grant No.4202073)

\bibliography{mybibfile}

\end{document}